\documentclass{article}
\usepackage[table]{xcolor}
\usepackage[preprint]{neurips_2024}

\usepackage[utf8]{inputenc} 
\usepackage[T1]{fontenc}    
\usepackage{hyperref}       
\usepackage{url}            
\usepackage{booktabs}       
\usepackage{amsfonts}       
\usepackage{nicefrac}       
\usepackage{microtype}      
\usepackage{lipsum}		
\usepackage{graphicx}
\usepackage{natbib}
\usepackage{doi}
\usepackage{multirow}
\usepackage{geometry}
\usepackage{amsmath}
\usepackage[most]{tcolorbox}

\title{URPO: A Unified Reward \& Policy Optimization Framework for Large Language Models}

\author{
    Songshuo Lu  \; \; 
    Hua Wang \; \; 
    Zhi Chen \; \; 
    Yaohua Tang \; \;   
    \\ \\Moore Threads AI \;
     \\
    \texttt{tangyaohua28@gmail.com}
}

\begin{document}
\maketitle

\begin{abstract}

Large-scale alignment pipelines typically pair a policy model with a separately trained reward model whose parameters remain frozen during reinforcement learning (RL). This separation creates a complex, resource-intensive pipeline and suffers from a performance ceiling due to a static reward signal. We propose a novel framework, Unified Reward \& Policy Optimization (URPO), that unifies instruction-following ("player") and reward modeling ("referee") within a single model and a single training phase. Our method recasts all alignment data—including preference pairs, verifiable reasoning, and open-ended instructions—into a unified generative format optimized by a single Group-Relative Policy Optimization (GRPO) loop. This enables the model to learn from ground-truth preferences and verifiable logic while simultaneously generating its own rewards for open-ended tasks.
Experiments on the Qwen2.5-7B model demonstrate URPO's superiority. Our unified model significantly outperforms a strong baseline using a separate generative reward model, boosting the instruction-following score on AlpacaEval from 42.24 to 44.84 and the composite reasoning average from 32.66 to 35.66. Furthermore, URPO cultivates a superior internal evaluator as a byproduct of training, achieving a RewardBench score of 85.15 and surpassing the dedicated reward model it replaces (83.55). By eliminating the need for a separate reward model and fostering a co-evolutionary dynamic between generation and evaluation, URPO presents a simpler, more efficient, and more effective path towards robustly aligned language models.

\end{abstract}

\section{Introduction}
The advent of Large Language Models (LLMs) such as GPT-4~\citep{achiam2023gpt4} and Llama~\citep{touvron2023llama} has revolutionized the field of artificial intelligence, demonstrating remarkable capabilities in understanding and generating human-like text. Post-training alignment, a process that refines the model's outputs to better align with specific instructions and human preferences, emerges to be a key technique for further enhancing the capabilities of LLMs. During this process, RLHF~\citep{christiano2017deep, ziegler2019fine} is adopted as a primary approach to align the outputs of LLMs~\citep{ouyang2022training}. The conventional RLHF pipeline is typically composed of two decoupled steps, where a reward model (the ``referee'') is first trained in a supervised manner with direct human feedback, and then this model is taken as a static reward function to optimize a policy model (the ``player'') through an optimization algorithm, such as proximal policy optimization (PPO)~\citep{schulman2017proximalpolicyoptimizationalgorithms}. However, the multistage and multimodel nature of the RLHF training pipeline poses several significant challenges.

First, it must maintain multiple models with different training recipes, which can lead to catastrophic forgetting problems and make the training pipeline cumbersome and inefficient. Moreover, since DeepSeek R1~\citep{guo2025deepseek}, GRPO~\citep{shao2024deepseekmath} training on rule-verifiable reasoning data has been commonly added to the post-training process, further complicating the pipeline. 

Second, the reward model is static and remains frozen after it is trained, but the policy model continuously improves during reinforcement learning. This leads to a phenomenon that we refer to as ``competence mismatch'', since an evolving, ever stronger player is judged by a static, unchanging referee. As the policy model explores more nuanced and complex responses, the fixed reward model may fail to provide accurate and informative feedback, potentially stifling the model's ultimate potential.

Third, this training pipeline results in data silos, where instruction-following data and preference data are exclusively used by the policy model and the reward model, respectively, thus impeding potential synergies between these two complementary skills.

To address these challenges, a natural question arises: \textit{Is it possible to redesign the RLHF training paradigm via a unified model that acts as both the player and the referee?} Arguably, this is possible if one could design a unified framework and train a single model to fulfill both instruction-following and self-evaluation tasks. The very single model learns not only how to accurately answer questions but also how to critically assess the quality of its own (and other) answers, establishing a cycle of self-improvement. It then enables generation and evaluation capabilities to co-evolve and mutually reinforce each other during the unified training loop.

\begin{figure}[t]
  \centering
  \includegraphics[width=\textwidth]{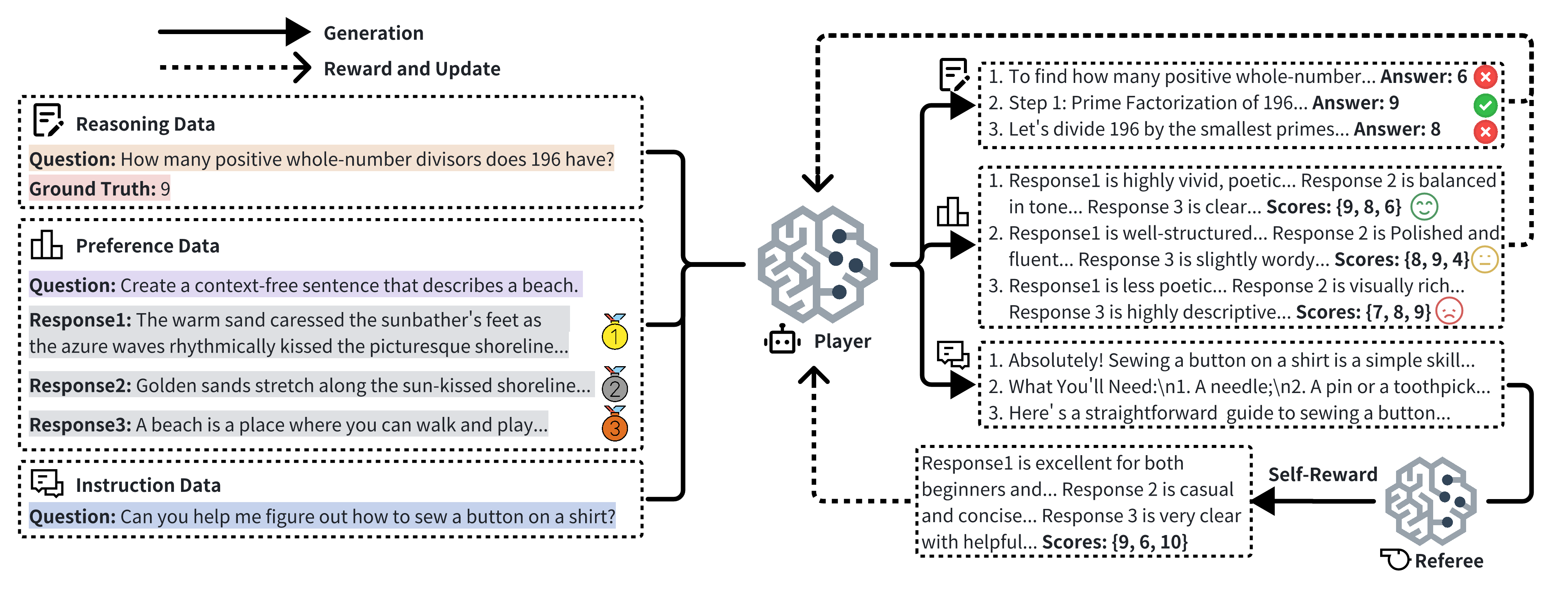}
  \caption{\textbf{URPO training loop.}
           A single large language model simultaneously plays the roles of
           \emph{player} (Policy Model) and \emph{referee} (Reward Model).
           Each mixed training batch contains three data types:
           (a) rule-verifiable reasoning problems with ground-truth answers,
           (b) preference triples with N-way ranking,
           and (c) open-ended instructions.
           The model first generates candidate responses, then evaluates them with either ground-truth checks or self-assessed ranking rewards before refining its parameters through GRPO, 
           allowing its generation and evaluation capabilities to co-evolve.}
  \label{fig:urpo_framework}
\end{figure}

To achieve this, we propose Unified Reward \& Policy Optimization (\textbf{URPO};
see Figure~\ref{fig:urpo_framework}), a novel training methodology that unifies three distinct data types, verifiable reasoning problems (e.g., math, code), open-ended instructions, and preference data, into a single training batch that fits the GRPO training. We recast preference triples in reward modeling into an N-way ranking prompt, reward the model with Kendall’s $\tau$ correlation against ground-truth orderings, and interleave rule-verifiable reasoning tasks to ground the policy in factual correctness. For each open-ended instruction, the model samples k roll-outs in GRPO training, combines them into a single reward question and prompts itself to rank them, and feeds the induced group rewards to GRPO training. This effectively helps us transform the entire post-training into a single cohesive GRPO training process.

Our contributions are threefold:

\begin{itemize}
\item \textbf{A Simpler and More Effective Unified Alignment Framework.} We propose URPO, an elegant single-model, single-process framework that replaces the standard reliance on a separate reward model for reinforcement learning alignment. Our experiments show this unified approach is not just simpler but also more effective. On Qwen2.5-7B, URPO significantly outperforms the baseline guided by a separately trained generative reward model, boosting the AlpacaEval score from 42.24 to \textbf{44.84} and the composite reasoning average from 32.66 to \textbf{35.66}.

\item \textbf{A Smarter, Self-Taught Referee.} With URPO, the model learns to judge its own answers while it learns to generate them. This process creates an internal "referee" that gets smarter over time, overcoming the limitations of a fixed, pre-trained reward model. In fact, our model became a better judge than a specialized reward model that was trained separately for the same task, achieving a higher score on the RewardBench benchmark (\textbf{85.15} vs. 83.55).

\item \textbf{Holistic Improvement Through Data Synergy.} We provide empirical proof that a unified data approach is critical. Our ablation studies show that a balanced data mixture is essential for creating a well-rounded model. 

\end{itemize}



\section{Related Work}
\subsection{Reinforcement Learning from Human (and AI) Feedback}
Early works demonstrate the promise of RLHF by fine-tuning language models with a learned reward function. \citet{ziegler2019fine} and \citet{stiennon2020learning} apply RLHF to tasks like summarization, training a reward model on human rankings and then optimizing a policy via PPO. This paradigm is popularized by InstructGPT~\citep{ouyang2022training}, which shows that a 1.3B model fine-tuned with human feedback could outperform a 175B GPT-3 on following instructions. The standard RLHF recipe involves a SFT step, then training a separate reward model on preference data, and finally optimizing the policy against this reward via PPO. While effective in aligning models with human intent, this multi-stage process is resource-intensive and delicate. Researchers have noted that RLHF training is "complex and often unstable", requiring careful reward modeling and constraint (e.g. KL penalties) to prevent the policy from drifting off-distribution. Recent methods therefore seek to simplify or improve this pipeline. For example, \citet{rafailov2023direct} propose Direct Preference Optimization (DPO), which forgoes explicit reward model training by re-parameterizing the objective so that the optimal policy can be solved via a simple classification-style loss on preference data. DPO eliminates the need for sampling-based RL updates and achieves alignment performance on par with PPO-based RLHF while being much easier to train. Another direction is Rejection Sampling Fine-Tuning (RFT), which simply fine-tunes on higher-ranked responses (selected via a reward model or heuristic) rather than running RL - a strategy shown to effectively incorporate preference signals with less complexity. These approaches retain the use of human preference data but streamline the optimization process.

A complementary thread replaces costly human feedback with AI-generated feedback. Anthropic's Constitutional AI~\citep{bai2022constitutional} pioneer the idea of using an AI model to critique and refine outputs according to a set of principles, thereby aligning a model to be harmless without direct human labels. More generally, Reinforcement Learning from AI Feedback (RLAIF) trains the reward model on preferences generated by an auxiliary large language model instead of humans. \citet{lee2023rlaif} demonstrate that RLAIF can achieve comparable results to RLHF on tasks like summarization and dialogue, essentially using one LLM as a labeler for another. Notably, they introduce a direct RLAIF variant that skips reward-model training entirely by querying an off-the-shelf LLM for rewards on-the-fly during policy optimization. This direct approach yielded even better performance than the standard two-model RLAIF pipeline. These results suggest that AI evaluators can substitute for human preference annotators, but they still treat the "judge" as a fixed external module (either a frozen reward model or a separate teacher LLM). In summary, contemporary preference-optimization methods like DPO and RLAIF ease the engineering burden of RLHF, yet they generally maintain a separation between the policy model and the source of reward signals. Our work instead explores fully merging the reward and policy roles within one model.

\subsection{Self-Critique and Self-Alignment Paradigms}
There is a parallel trend toward self-training and self-critiquing paradigms, where the model itself plays the dual roles of content generator and evaluator. Several recent works show that an LLM can improve its performance by generating feedback or critiques of its own outputs and learning from them. For example, in the area of factual accuracy, \citet{zhang2024self} leverage a model’s own knowledge to detect hallucinations: they prompt the LLM to self-evaluate the factual correctness of its answer, and use those self-generated assessments as labels to fine-tune the model via DPO. This self-alignment for factuality significantly reduces hallucinations without requiring human annotations. Another line of work has explored iterative refinement: \citet{madaan2023self} enable an LLM to iteratively critique and revise its responses (e.g. the Self-Refine method), while \citet{shinn2023reflexion} and others propose frameworks for an AI agent to reflect on wrong answers and correct itself in multi-turn interactions. These approaches often operate at inference-time or require an external oracle, but they hint at the potential of models to be their own "teachers".

Recently, more formal training frameworks for self-rewarding have been proposed. \citet{zhao2025learning} and \cite{zhou2025reinforcing} use a LLM's own confidence/probability of generating the reference answer as its sole reward signal. \citet{yuan2025selfrewardinglanguagemodels} introduce an iterative self-rewarding algorithm in which a single LLM is used both as the "actor" (producing responses) and as the "judge" (evaluating these responses). In each round, the model generates multiple candidate answers to a prompt and then, via an LLM-as-a-Judge prompt, assigns scores or preferences to its own candidates. The model is then fine-tuned to prefer the candidates it judged as better, and this process repeats. \citet{wu2024meta} point out that one limitation of this approach is that the model’s judgment ability may lag behind its generation ability, causing training to plateau. They propose a "meta-rewarding" extension wherein the model also evaluates its own evaluations: a meta-judge role that critiques the judge’s decisions, providing an extra training signal to improve the model’s evaluative skill. \citet{yuan2025selfrewardinglanguagemodels} emphasize enhancing instruction-following capabilities without concurrent improvement in reward modeling while \citet{wu2024meta} recognize the necessity for simultaneous advancement in reward modeling but relies on self-generated model data for this enhancement. Our proposed URPO posits that effective development of reward modeling capabilities requires explicit training on external human-annotated preference data. This external grounding ensures rewards stay objective and independent from the model’s biases. Without external data, models might exploit weaknesses in their own reward systems ("reward hacking") rather than learning meaningful alignment. The reliance solely on instruction-following datasets is insufficient for authentic reward model improvement.

\section{Methodology}

Traditional large language model (LLM) alignment pipelines often decouple generative capabilities (the ``player'') from evaluative mechanisms (the ``referee''). This separation introduces complexities, including the management of multiple models, risks of catastrophic forgetting during sequential training stages, and a static reward signal. This section presents URPO, a novel method that addresses these challenges by training a single language model with excellence in both instruction-following and self-evaluation. This holistic approach unifies disparate post-training stages, such as verifiable reasoning, open-ended generation, and preference alignment, into a single, cohesive GRPO process, fostering a continuous, self-reinforcing improvement loop.

\subsection{Preliminaries: Decoupled Reinforcement Learning Alignment}
\label{sec:preliminaries}

After an initial Supervised Fine-Tuning (SFT) phase, the alignment of modern large language models heavily relies on reinforcement learning. The standard paradigm, however, follows a decoupled approach, employing separate mechanisms to handle tasks with verifiable answers (e.g., math, code) versus those requiring alignment with open-ended human preferences.

For open-ended tasks, the conventional RLHF pipeline consists of two distinct stages. First, a separate \textbf{Reward Model ($r_\phi$)} is trained on a large dataset of human preference pairs, learning to assign a scalar score that predicts human judgment. Subsequently, this reward model is "frozen" to act as a static referee, providing a reward signal to fine-tune the main language model, the \textbf{Policy Model ($\pi_\theta$)}.

A powerful algorithm often used to drive both types of alignment is GRPO~\citep{shao2024deepseekmath}. Its core idea is to sample a group of $G$ responses $\{o_i\}_{i=1}^G$ from the policy and calculate a value-free, group-relative advantage $\hat{A}_i$ for each response based on its reward $R_i$ compared to the group average. The policy is then updated using the canonical GRPO objective:

\begin{equation*}
\small
\begin{aligned}
\mathcal{J}_\text{GRPO}(\theta)& = \mathbb{E}_{(q,a)\sim \mathcal{D}, \{o_i\}_{i=1}^G\sim \pi_{\theta_\text{old}}(\cdot\mid q)} \\&
\Bigg[ \frac{1}{G}\sum_{i=1}^{G} \underbrace{\frac{1}{|o_i|}}_{\text{Length Norm.}} \sum_{t=1}^{|o_i|} \Bigg(
\underbrace{\min \Big( r_{i,t}(\theta) \hat{A}_{i}, 
\ \text{clip} \Big( r_{i,t}(\theta), 1 - \varepsilon, 1 + \varepsilon \Big) \hat{A}_{i} \Big)}_{\text{Clipped Surrogate Objective}}
- \underbrace{\beta D_{\text{KL}}(\pi_{\theta} || \pi_{\text{ref}})}_{\text{KL Penalty}}
\Bigg) \Bigg].
\label{eq:canonical_grpo}
\end{aligned}
\end{equation*}

This objective maximizes the advantage $\hat{A}_i$ weighted by an importance ratio $r_{i,t}(\theta)$. The standard formulation typically includes a per-response length normalization ($1/|o_i|$), a clipping mechanism ($\varepsilon$) to stabilize the update, and a KL divergence penalty ($\beta$) to regularize the policy against a reference model $\pi_\text{ref}$.

Crucially, in this standard paradigm, the system is fundamentally decoupled. The "player" $\pi_\theta$ and the "referee" $r_\phi$ are trained and operate as independent entities. 
The policy for reasoning is optimized against hard-coded rules, while the policy for instruction-following is optimized against a separate, static reward model. This separation leads directly to the shortcomings we identified: a fragmented pipeline, a static competence ceiling for the reward signal, and a lack of internal consistency. These pronounced limitations motivate a fundamental rethinking of the alignment process, leading us to our proposed solution.

\subsection{URPO}
The core innovation of URPO lies in its ability to reformat all training instances into a structure amenable to direct optimization by GRPO, thereby enabling a single model to simultaneously learn generative and evaluative competencies. This approach integrates instruction-following and self-rewarding capabilities within a single LLM, allowing it to learn from verifiable reasoning tasks, open-ended generation, and preference data concurrently.




\subsubsection{Formulating Reward Modeling as a Generative GRPO Task}
\label{sec:rm}
The paradigm for training reward models has evolved beyond simple scoring mechanisms. Traditionally, reward models have been trained as separate classifiers or regressors on human preference data (e.g., $\langle q, a_{\text{chosen}}, a_{\text{reject}} \rangle$), learning to output a scalar score, often optimized with a ranking loss like the Bradley-Terry model or BPR loss~\citep{rendle2012bpr}. While effective, this approach creates a static referee, decoupled from the policy model it aims to guide.
More recently, the landscape has diversified with the advent of LLM-as-a-Judge~\citep{zheng2023judging}. This approach leverages a powerful, external large language model (often a proprietary model like GPT-4) to evaluate and score responses. While demonstrating strong correlation with human judgments, this method introduces significant dependencies, either incurring high operational costs through API calls or requiring the fine-tuning of a powerful open-source model on vast amounts of costly, detailed critique data to act as a stand-in judge. Furthermore, emerging research explores training reward models themselves through reinforcement learning paradigms~\citep{liu2025inference}. However, the optimal methods for integrating these dynamically trained referees back into a stable downstream policy-training pipeline are still an active area of investigation.

URPO fundamentally transforms this landscape by sidestepping the need for any separate or external reward model. It recasts reward modeling as an intrinsic generative task performed by the primary policy model itself. This eliminates the complexities of managing a second model, the costs associated with external judges, and the uncertainties of novel reward training schemes, creating a fully self-contained and co-evolutionary system.

Our framework constructs input prompts by concatenating the preference data into a multi-response evaluation template, instructing the model to act as an evaluator:
\begin{tcolorbox}[colback=gray!10,enhanced,sharp corners,frame hidden,breakable]
You are a skilled expert at scoring responses. You will be given a question with multiple responses. Evaluate and score each response. \\
\texttt{[The Begin of Question]\{$q$\}[The End of Question]}\\
\texttt{[The Begin of Response 1]\{$y_w$\}[The End of Response 1]}\\
...\\
\texttt{[The Begin of Response N]\{$y_l$\}[The End of Response N]}\\
\#\#\#\# Output Format Requirements \#\#\#\#\\
Scores: <the overall comprehensive score of all responses in order, separate by comma in the boxed, e.g., \textbackslash boxed\{x, x\} if there exists 2 responeses>.
\end{tcolorbox}

In real model training, the number of reference samples is not restricted to two but can be actually many, provided that a ground-truth-based rank ordering exists. 
We integrate these preference data into the training batch with the aforementioned prompt through the following GRPO steps:
\begin{itemize}
    \item \textbf{Rollout}: For a given question $q$ and a set of $N$ response candidates $\{a_i\}_{i=1}^N$ (derived from preference data), the policy $\pi_{\theta_{\text{old}}}$ generates a group of $G$ evaluation outputs, each containing predicted scores for the responses.
    \item \textbf{Reward}: Each generated evaluation output provides a predicted ranking of the candidate responses based on the assigned scores. Let $\mathbf{s}=(s_1,\dots,s_N)$ be the predicted scores for responses $\{a_i\}_{i=1}^N$, yielding a predicted ranking vector $\mathbf{r}$. We compare this predicted ranking with the ground-truth ranking $\mathbf{g}$ (derived from human/AI preferences) using Kendall's $\tau$ correlation coefficient as the scalar reward signal:
\[
R(q, \{a_i\}_{i=1}^N)=\tau_K(\mathbf r,\mathbf g)
     =\frac{2}{N(N-1)}
     \sum_{1\le i<j\le N}
     \operatorname{sign}\!\bigl[(s_i-s_j)(g_i-g_j)\bigr],
\]
where $\operatorname{sign}(x)\in\{-1,0,1\}$ is the sign function. This reward $R(q)\in[-1,1]$ is $1$ for perfect agreement and $-1$ for perfect disagreement.
    \item \textbf{GRPO Update}: These Kendall's $\tau$ rewards are then used within the GRPO objective to update the policy, encouraging it to generate evaluations that align more closely with expert preferences.
\end{itemize}

\subsubsection{Self-Rewarding for Open-Ended Generation}
A critical innovation of URPO is its capacity for self-evaluation, which eliminates the need for an external reward model. This process dynamically integrates the model's generative and evaluative roles within the RL loop. For an open-ended prompt q, the policy $\pi_{\theta}$ first acts as a "player", generating a set of $G$ diverse candidate responses$\{o_i\}_{i=1}^G$ in the rollout phase.

The $G$ generated responses are formatted into a single, comprehensive evaluation prompt (as described in Section \ref{sec:rm}), and the same policy $\pi_{\theta}$ performs a forward pass to produce a text containing self-assigned scores for all $G$ candidates (e.g., \texttt{Scores: $[score_1, score_2, \dots, score_G]$}). These extracted scores are then directly used as the rewards for their corresponding responses in the GRPO update. This closes the self-improvement loop, allowing the model to refine its generative capabilities based on its own evolving critical judgment, without any external supervision on the reward signal itself.

\begin{table}[b]
\centering
\caption{Task Unification in the URPO Framework. All tasks are optimized via a single GRPO objective, but differ in the primary model role being trained and the source of the reward signal.}
\label{table:tasks}
\begin{tabular}{@{}lllc@{}}
\toprule
\textbf{Task Category} & \textbf{Input Format} & \textbf{Primary Model Role} & \textbf{Reward Signal Source} \\
\midrule
Verifiable Reasoning & $\langle q, y_{gt} \rangle$ & Player (Generator) & Rule-based Verifier \\
Open-Ended Generation & $\langle q \rangle$ & Player (Generator) & Self-Generated Scores \\
Preference Alignment & $\langle q, \{y_i\}_{\text{ranked}} \rangle$ & Referee (Evaluator) & Kendall's $\tau$ vs. Ground-Truth \\
\bottomrule
\end{tabular}
\end{table}

\subsubsection{Co-evolution of Player and Referee}
URPO's strength lies in its unified GRPO-based training loop, which processes mixed batches encompassing all three data types: verifiable reasoning, open-ended generation, and preference data. 
As illustrated in Table~\ref{table:tasks}, while all tasks are optimized using the same underlying algorithm, our integrated approach fosters a profound synergistic relationship between the model's generative ("player") and evaluative ("referee") abilities. Incorporating reasoning data enhances the model's logical coherence and problem-solving skills, which in turn improves its capacity for discerning reward scoring by enabling more sophisticated comparative analysis of candidate responses.

Instead, a more astute "referee" provides a more accurate and challenging reward signal, effectively guiding the "player" towards generating increasingly superior and nuanced responses. This continuous, co-evolutionary dynamic, encapsulated within a single model and a unified training process, leads to more robust and efficient self-improvement, transcending the limitations of static, decoupled alignment pipelines. Furthermore, this architecture significantly reduces operational overhead, requiring only one model in memory and a single training job.
As illustrated in Table~\ref{table:tasks}, while all tasks are optimized using the same underlying algorithm, they strategically differ in how they leverage the model's dual roles as a "player" and "referee" and in what constitutes their reward signal.

\section{Experiments and Evaluation}

We conduct a series of experiments to empirically validate the effectiveness of URPO. Our primary objective is to demonstrate that a single, unified training process can simultaneously enhance a model's capabilities in three distinct domains: general instruction-following, complex logical reasoning, and nuanced reward evaluation. Our analysis compares URPO against established post-training pipelines and investigates its performance across multiple base models and data compositions.

\subsection{Experiment Setting}

Our primary experiments are conducted on Qwen2.5-7B-Base~\citep{qwen2.5} to ensure a fair comparison with existing methods. To demonstrate the generalizability of our approach, we also extend our analysis to Qwen3\citep{qwen3} and Llama3.1~\citep{grattafiori2024llama} series models.

\subsubsection{Baselines} 
\label{sec:baselines}
We compare URPO against two strong post-training baselines. 
\begin{itemize}
    \item \textbf{SFT + GRPO.} A standard multi-stage alignment approach. In this pipeline, the base model first undergoes SFT on a general-purpose, open-source instruction dataset to endow it with foundational instruction-following capabilities. Subsequently, this SFT model is trained using GRPO on our target data mix (reasoning and open-ended instructions).

    \item \textbf{Direct GRPO.} A pipeline bypasses the initial SFT stage and applies the GRPO process directly to the base model.
\end{itemize}

For both baselines, the GRPO stage requires an external reward signal for open-ended instructions. To this end, we trained two types of reward models (RMs) for a robust comparison. 

\begin{itemize}
    \item \textbf{RM-score.} A traditional scoring-based reward model architecture. This model takes a prompt-response pair as input and outputs a single scalar value representing its quality.

    \item \textbf{RM-gen.} A generation-based reward model trained to emulate the evaluative task format described in Section~\ref{sec:rm}. This model is prompted with multiple candidate responses and generates a text containing scores for each, providing a stronger and more comparable baseline to our URPO method.
\end{itemize}

    

    

\subsubsection{Training Data} 

Our experiments leverage a diverse mix of publicly available datasets:

\textbf{Reasoning Data}: We use Skywork-OR1-RL-Data~\citep{he2025skywork}, which contains approximately 105k mathematical problems and 14k coding challenges.
    
\textbf{Preference Data}: To train our baseline RMs and for the preference alignment component of URPO, we amalgamate five datasets: HelpSteer3~\citep{wang2025helpsteer3preferenceopenhumanannotatedpreference}, UltraFeedback~\citep{cui2023ultrafeedback}, Skywork-Reward-Preference~\citep{liu2024skywork}, Nectar~\citep{starling2023}, and offsetbias~\citep{park2024offsetbias}.

\textbf{Instruction Data}: For open-ended instruction following, we use 180k prompts from the prompt-collection-v0.1\footnote{https://huggingface.co/datasets/OpenRLHF/prompt-collection-v0.1} dataset.


\subsubsection{Evaluation Benchmarks} 

We assess model capabilities across three axes:

\textbf{Reasoning}: We use a challenging suite of math benchmarks, including GSM8K~\citep{cobbe2021training}, MATH-500~\citep{hendrycks2021measuring}, and recent competition problems from AIME (2024, 2025) and HMMT (2024, 2025).
    
\textbf{Instruction Following}: We use AlpacaEval 2.0~\citep{li2023alpacaeval}, with gpt-4o-2024-08-06\footnote{https://platform.openai.com/docs/models/gpt-4o} serving as the judge for robust evaluation.

\textbf{Reward Modeling}: We evaluate the emergent reward modeling capabilities using RewardBench~\citep{lambert2024rewardbench} and RMB~\citep{zhou2024rmb}.

\subsubsection{Implementation Details}
All experiments are conducted using the \texttt{VERL}~\citep{sheng2025hybridflow} framework. We adopt several advanced training techniques to ensure stability and mitigate common RL biases.

\textbf{GRPO Hyperparameters and Loss Formulation.} 
We use the AdamW optimizer~\citep{loshchilov2017decoupled} with a learning rate of $5 \times 10^{-7}$ and a cosine decay schedule. The global batch size is 256 prompts, with $G=8$ responses sampled per prompt at a temperature of $1.0$ and a maximum generation length of 4096 tokens. The policy is updated for $4$ epochs on each collected batch. To refine the training objective, we incorporate two critical modifications from DAPO~\citep{yu2025dapo}. First, to mitigate length bias, we remove the per-response length normalization term ($1 / |o_i|$) from the loss calculation. Second, we employ an asymmetric "clip\_higher" strategy, restricting the importance sampling ratio $r_{i,t}(\theta)$ to the range $[0.8, 1.28]$ to prevent policy collapse. Furthermore, to maximize exploration given the base model's initial limitations, we completely remove the KL-divergence constraint by setting its coefficient $\beta$ to 0.


\textbf{Two-Stage Curriculum Strategy.}
To ensure the model first develops reliable evaluative capabilities before applying them to subjective tasks, we adopt a two-stage curriculum learning strategy:
\begin{itemize}
    \item \textbf{Phase 1: Warmup.} For the initial 100 training steps, the model is trained exclusively on a mixture of verifiable reasoning data and preference data. The objective of this phase is to cultivate robust reasoning and evaluation faculties by using tasks with clear, objective reward signals (i.e., rule-based verification and ground-truth preference ranks). This effectively bootstraps the model's ability to act as a proficient "referee".
    \item \textbf{Phase 2: Unified Training.} After the warmup phase, the open-ended instruction data is introduced into the training mix. At this stage, the model, now equipped with a more reliable judgment faculty, can generate higher-quality self-rewards for the instruction-following task. This creates a more effective and stable self-improvement loop, allowing all three capabilities to co-evolve synergistically for the remainder of the training process.
\end{itemize}

\subsection{Results and Analysis}

\subsubsection{URPO Outperforms Standard RLHF Pipelines}

Our primary results, presented in Table~\ref{table:basecomparision}, compare URPO against several strong baselines in two distinct experimental settings. These baselines consist of standard GRPO pipelines guided by either a separately trained scoring reward model (RM-score) or a generative one (RM-gen), as detailed in Section~\ref{sec:baselines}.

When applied directly to the Qwen2.5-7B-Base model, our URPO pipeline is the unequivocal top performer. It achieves the highest scores across all metrics, culminating in a commanding lead on the composite reasoning average of 35.66. This result highlights URPO's strength as a comprehensive, end-to-end alignment solution that significantly surpasses standard baselines.

The second experimental group begins from an SFT checkpoint, which we created by supervised fine-tuning the base model on the TULU 3 dataset\footnote{https://huggingface.co/datasets/allenai/tulu-3-sft-mixture}. The SFT process itself creates a specialized but brittle model: its reasoning average improves from 24.73 to 26.57, but its general instruction-following ability collapses (AlpacaEval: 14.53). This performance drop may be attributed to an imbalanced data distribution in the fine-tuning mixture. However, our primary focus is to compare the relative performance gains each alignment method brings to this common starting point. When applied to this compromised checkpoint, URPO delivers the most effective recovery, restoring instruction-following capability to a group-best 42.36 on AlpacaEval and achieving the highest overall reasoning average of 31.83.


Finally, the results from both experimental settings, when viewed together, highlight the robust versatility of our URPO framework. It not only provides the most effective end-to-end alignment solution when applied directly to a base model, but it also delivers the largest performance gains when used as a second-stage alignment method on an SFT checkpoint. This demonstrates that regardless of the starting point, our unified, co-evolutionary training approach consistently provides the most effective path to holistically improving a model's instruction-following and reasoning capabilities.

\begin{table}[t]
    \centering
    \footnotesize
    \caption{Main performance comparison of URPO against baseline alignment pipelines on Qwen2.5-7B. Results are grouped by the starting checkpoint (Base vs. SFT). URPO consistently achieves the strongest overall performance in both settings.}
    \begin{tabular}{@{}l c|*{6}{c} c@{}}
        \toprule
        \multirow{2}{*}{\textbf{Pipeline}} & \textbf{Alpaca} & \multirow{2}{*}{\textbf{GSM8K}} & \textbf{MATH} & \textbf{AIME} & \textbf{AIME} & \textbf{HMMT} & \textbf{HMMT} & \textbf{Math} \\
         & \textbf{Eval} & & \textbf{500} & \textbf{2024} & \textbf{2025} & \textbf{2024} & \textbf{2025} & \textbf{Avg.} \\
        \midrule
        \multicolumn{9}{@{}l}{\textit{Starting from Qwen2.5-7B Base}} \\
        \quad Base Model & 41.61 & 82.71 & 59.40 & 3.33 & 0.83 & 2.08 & 0.00 & 24.73 \\
        \quad Instruct Baseline\textsuperscript{\textdagger} & 43.11 & 91.51 & 72.20 & 6.25 & 5.83 & 5.83 & 0.42 & 30.34 \\
        \quad + GRPO (RM-score) & 36.40 & 90.37 & 74.60 & 11.25 & 7.92 & 5.83 & 2.50 & 32.08 \\
        \quad + GRPO (RM-gen) & 42.24 & 90.75 & 74.80 & 13.33 & 10.83 & 4.58 & 1.67 & 32.66 \\
        \quad + URPO & \textbf{44.84} & \textbf{91.96} & \textbf{77.40} & \textbf{17.08} & \textbf{15.83} & \textbf{8.33} & \textbf{3.33} & \textbf{35.66} \\        
        \midrule
        \multicolumn{9}{@{}l}{\textit{Starting from Qwen2.5-7B+SFT}} \\
        \quad SFT Checkpoint & 14.53 & 83.02 & 71.00 & 2.08 & 1.67 & 1.25 & 0.42 & 26.57 \\
        \quad + GRPO (RM-score) & 28.32 & 87.72 & 71.00 & 11.25 & 5.42 & 4.58 & 2.08 & 30.34 \\
        \quad + GRPO (RM-gen) & 39.38 & \textbf{90.37} & 72.80 & 11.25 & 7.08 & \textbf{5.83} & \textbf{2.50} & 31.64 \\
        \quad + URPO & \textbf{42.36} & 87.79 & \textbf{74.00} & \textbf{12.50} & \textbf{9.58} & 5.00 & 2.08 & \textbf{31.83} \\
        \bottomrule
    \end{tabular}
    \par\addvspace{2pt}
    \footnotesize{\textsuperscript{\textdagger}Qwen2.5-7B-Instruct model, representing a standard SFT+RLHF pipeline for comparison.}
    \label{table:basecomparision}
\end{table}

\subsubsection{Generalizability and Base Model Sensitivity}

To verify that URPO's benefits are not confined to a single model family, we tested its performance across diverse state-of-the-art base models, yielding insights into both the framework's generalizability and the preconditions for successful alignment.

As shown in Table~\ref{table:variousmodels}, URPO demonstrates robust performance gains across different models. On the advanced Qwen3 model, our unified method achieved the best performance on a majority of benchmarks, including the highest scores for instruction following (AlpacaEval 39.25) and top-tier results across the reasoning suite. This advantage was even more decisive on the Llama3.1-Base model, where URPO established itself as the top-performing method across nearly all metrics, significantly boosting both instruction following (AlpacaEval to 41.49) and the composite reasoning average to 27.43.


It is critical to note, however, that these successful results on the Llama3.1 architecture were achieved after an initial experimental setback. In line with challenges noted across the community~\citep{gandhi2025cognitive}, applying reinforcement learning directly to the vanilla Llama3.1-8B model resulted in training instability and performance collapse. This prompted us to switch to \texttt{OctoThinker}~\citep{wang2025octothinker}, a version of Llama3.1 specifically fortified with extensive "mid-training" on high-quality reasoning data. 
This experience provides a powerful insight: the success of applying intensive reinforcement learning directly to a base model is highly conditional on that model already possessing a foundational "seed" of competence.


\begin{table}[htbp]
    \centering
    \footnotesize
    \caption{URPO Performance Comparison Across Various Base Models.}
    \begin{tabular}{@{}l c|*{6}{c} c@{}} 
        \toprule
        \multirow{2}{*}{\textbf{Pipeline}} & \textbf{Alpaca} & \multirow{2}{*}{\textbf{GSM8K}} & \textbf{MATH} & \textbf{AIME} & \textbf{AIME} & \textbf{HMMT} & \textbf{HMMT} & \textbf{Math} \\
         & \textbf{Eval} & & \textbf{500} & \textbf{2024} & \textbf{2025} & \textbf{2024} & \textbf{2025} & \textbf{Avg.} \\
        \midrule
        \multicolumn{9}{@{}l}{\textit{Qwen3-8B-Base Series}} \\
        \quad Qwen3 Base & 20.75 & 59.89 & 38.80 & 2.08 & 2.92 & \textbf{8.75} & 0.42 & 18.81 \\
        \quad + RM-gen & 36.77 & 93.40 & 83.00 & \textbf{26.25} & 18.75 & \textbf{8.75} & 8.33 & \textbf{39.75} \\
        \quad + URPO & \textbf{39.25} & \textbf{94.01} & \textbf{83.40} & 23.33 & \textbf{19.17} & 8.33 & \textbf{9.17} & 39.57 \\
        \midrule
        \multicolumn{9}{@{}l}{\textit{Llama3.1-8B-Base Series}} \\
        \quad Llama3.1 Base* & 28.70 & 65.88 & 44.00 & 0.83 & 1.67 & 0.83 & 0.00 & 18.87 \\
        \quad + RM-gen & 36.15 & 86.05 & 61.20 & 6.67 & 1.25 & \textbf{2.92} & 0.42 & 26.42 \\
        \quad + URPO & \textbf{41.49} & \textbf{87.04} & \textbf{64.20} & \textbf{7.08} & \textbf{2.92} & 2.50 & \textbf{0.83} & \textbf{27.43} \\
        \bottomrule
    \end{tabular}
    \par\addvspace{2pt}
    \footnotesize{*The Llama3.1-based model used is OctoThinker-8B-Hybrid-Base.}
    \label{table:variousmodels}
\end{table}

\begin{table}[htbp]
    \centering
    \footnotesize
    \tabcolsep=0.18cm
    \caption{Ablation study on data composition. A balanced mixture of Preference (P), Reasoning (R), and Instruction (I) data yields the best overall performance, highlighting the synergy between different data types. Grayed cells indicate significant performance degradation.}
    \begin{tabular}{@{}l c|*{6}{c} c@{}} 
        \toprule
        \multirow{2}{*}{\textbf{Data Mix Ratio (P:R:I)}} & \textbf{Alpaca} & \multirow{2}{*}{\textbf{GSM8K}} & \textbf{MATH} & \textbf{AIME} & \textbf{AIME} & \textbf{HMMT} & \textbf{HMMT} & \textbf{Math} \\
         & \textbf{Eval} & & \textbf{500} & \textbf{2024} & \textbf{2025} & \textbf{2024} & \textbf{2025} & \textbf{Avg.} \\
        \midrule
        1 : 1 : 0 (No Instruction) & \cellcolor{lightgray}31.43 & \textbf{92.87} & \textbf{79.00} & 13.33 & 15.41 & 6.67 & 2.91 & 35.03 \\
        1 : 0 : 1 (No Reasoning) & 43.11 & 72.63 & 66.80 & 12.91 & 12.08 & 3.75 & 3.33 & \cellcolor{lightgray}28.58 \\
        \midrule
        1 : 2 : 1 & 42.61 & 91.28 & 77.20 & 15.00 & 15.83 & 7.08 & 4.17 & 35.09 \\
        1 : 1 : 1 & \textbf{44.84} & 91.96 & 77.40 & 17.08 & 15.83 & 8.33 & 3.33 & 35.66 \\
        2 : 1 : 1 & 42.86 & 92.04 & 77.60 & 16.67 & 15.00 & 5.42 & 2.50 & 34.87 \\
        3 : 1 : 1 & 42.11 & 92.65 & 78.20 & \textbf{20.42} & 14.58 & 6.67 & 2.08 & \textbf{35.77} \\
        \bottomrule
    \end{tabular}
    \label{table:mixratioinstruct}
\end{table}

\subsubsection{Ablation Study: The Synergy of Data Composition}

To dissect the contribution of each data type within the URPO framework, we conducted a comprehensive ablation study on the Qwen2.5-7B-Base model. The results, presented in Table~\ref{table:mixratioinstruct} and Table~\ref{table:mixratioreward}, reveal a clear synergy between preference, reasoning, and instruction data, and validate our core hypotheses regarding the benefits of a unified training approach.


\textbf{Trade-offs in Generative Capabilities.} Table~\ref{table:mixratioinstruct} demonstrates that data composition directly governs the model's final generative skills. Excluding a key data type results in a pronounced trade-off. For instance, the \texttt{1:1:0} mixture (lacking instruction data) yields high reasoning scores (Math Avg. 35.03) but suffers a significant degradation in instruction-following performance (AlpacaEval 31.43). Conversely, the \texttt{1:0:1} mixture (lacking reasoning data) improves on AlpacaEval (43.11) at the cost of the lowest reasoning average (28.58). A balanced \texttt{1:1:1} mixture provides the best overall performance, achieving the highest AlpacaEval score while maintaining highly competitive reasoning capabilities. It is crucial to note that when comparing these results to our strongest baseline in Table~\ref{table:basecomparision}, the URPO framework demonstrates remarkable robustness. Our optimal \texttt{1:1:1} configuration decisively outperforms the "GRPO (RM-gen)" model, and nearly all mixture ratios yield a better overall balance of capabilities, validating the inherent superiority of the unified training approach regardless of the specific data composition.

\textbf{Impact of Data Composition on Evaluative Capabilities.} The analysis in Table~\ref{table:mixratioreward} establishes three key findings regarding the model's emergent evaluative skills:
\begin{itemize}

    \item \textbf{Preference data as a prerequisite for self-Rewarding.} The results first confirm the critical role of preference data. The \texttt{0:1:1} mixture, which lacks this component, demonstrates inadequate evaluative capabilities, resulting in a catastrophic RewardBench Mean score of 62.39. This finding has broader implications for self-rewarding reinforcement learning methodologies. It strongly suggests that, at least when starting from a base model, purely self-generated reward signals are insufficient to bootstrap a reliable internal evaluator. Human preference data appears to be an essential ingredient to initially ground and guide the model's judgment faculty before it can effectively self-improve.

    \item \textbf{Reasoning data enhances evaluative accuracy.} Our results empirically validate the central hypothesis that reasoning skills are transferable to evaluative tasks. This is most evident when comparing against our dedicated \texttt{RM-gen} baseline, which is trained exclusively on preference data (effectively a \texttt{1:0:0} mix). The \texttt{1:1:0} mixture (Preference + Reasoning) achieves a RewardBench Mean of 85.15, substantially surpassing the \texttt{RM-gen}'s score of 83.55. This demonstrates that integrating logical reasoning problems is a more effective method for improving a model's judgment than relying on preference data alone. This finding has significant implications beyond our unified framework: it suggests that even for the goal of training a powerful standalone reward model, incorporating reasoning data into the RL process is a beneficial strategy for enhancing its final evaluative accuracy.

    \item \textbf{Increasing preference data yields diminishing returns.} While increasing the preference data ratio from \texttt{1:1:0} to \texttt{2:1:0} provides a marginal improvement in the RewardBench Mean (from 85.15 to 85.31), the small gain suggests that a balanced approach is more data-efficient than simply scaling up preference data volume.
    
\end{itemize}

\begin{table}[t]
    \centering
    \footnotesize
    \caption{Analysis of emergent reward modeling capabilities on the RewardBench benchmark. Models are evaluated against baselines, including specialized reward models (RMs).}
    \begin{tabular}{@{}l*{5}{c}@{}}
        \toprule
        \textbf{Pipeline} & \textbf{Chat} & \textbf{Chat Hard} & \textbf{Safety} & \textbf{Reasoning} & \textbf{Mean} \\
        \midrule
        \multicolumn{6}{@{}l}{\textit{Baseline Models and Reward Models (RMs)}} \\
        \quad Qwen Base & 65.64 & 46.60 & 65.81 & 50.84 & 57.22 \\
        \quad Qwen-Instruct & 96.37 & 57.79 & 79.46 & 80.37 & 78.50 \\
        \quad RM-score (ours) & 94.41 & 72.81 & 68.51 & 82.97 & 79.68 \\
        \quad RM-gen (ours) & 96.65 & 63.82 & 86.89 & 86.84 & 83.55 \\
        \midrule
        \multicolumn{6}{@{}l}{\textit{URPO Data Mixture Ablation (Ratio P:R:I)}} \\
        \quad 0 : 1 : 1 & 76.40 & 48.90 & 61.35 & 62.91 & 62.39 \\
        \quad 1 : 0 : 1 & 93.16 & 71.71 & 88.65 & 85.30 & 84.70 \\
        \quad 1 : 1 : 0 & 94.69 & 73.68 & 87.36 & 84.88 & 85.15 \\
        \quad 2 : 1 : 0 & 95.81 & 71.82 & 85.95 & 87.67 & \textbf{85.31} \\
        \bottomrule
    \end{tabular}
    \label{table:mixratioreward}
\end{table}

\section{Conclusion}

In this work, we addressed the increasing fragmentation and performance ceilings of conventional, decoupled alignment pipelines. We introduced Unified Reward \& Policy Optimization (URPO), a novel framework that challenges this paradigm by training a single language model to simultaneously master generative ("player") and evaluative ("referee") capabilities. By unifying three distinct data types—verifiable reasoning, human preference, and open-ended instructions—into a single, cohesive GRPO process, URPO fosters a powerful co-evolutionary learning loop.


Our extensive experiments empirically validate the effectiveness of this unified approach. 
We demonstrated that URPO consistently outperforms standard GRPO baselines that depend on a separate, pre-trained reward model across diverse model families.
When applied directly to a base model, URPO achieved the highest performance on both instruction-following and a suite of challenging reasoning benchmarks, yielding a composite reasoning average of 35.66 on Qwen2.5-7B. Our ablation studies further revealed the mechanisms behind this success: a balanced mixture of preference, reasoning, and instruction data is critical for developing a well-rounded model, and integrating verifiable reasoning data is a potent strategy for enhancing a model's evaluative accuracy (e.g., 85.15 vs. 83.55 on RewardBench). This holistic enhancement is a key advantage over methods that apply reinforcement learning solely to data with verifiable rewards.
URPO's unified data strategy fosters a more comprehensive advancement across reasoning, instruction-following, and evaluative capabilities simultaneously.


Ultimately, URPO provides an elegant, effective, and efficient framework for holistic LLM alignment. By unifying the player and the referee, URPO not only simplifies the complex alignment process but also fosters models with greater internal consistency, breaking the performance ceiling imposed by static reward models. This work suggests that the future of LLM alignment may lie not in building more elaborate, fragmented pipelines, but in creating more integrated learning architectures that enable models to reason about, evaluate, and improve upon their own outputs in a continuous, self-reinforcing loop.


\bibliographystyle{plainnat}
\bibliography{references}

\newpage

\appendix


\end{document}